%% file: ijcai24.tex
\documentclass{article}
\usepackage{ijcai24}

\usepackage{times}
\usepackage{soul}
\usepackage{url}
\usepackage[hidelinks]{hyperref}
\usepackage[utf8]{inputenc}
\usepackage[small]{caption}
\usepackage{graphicx}
\usepackage{amsmath}
\usepackage{amsthm}
\usepackage{booktabs}
\usepackage{algorithm}
\usepackage[switch]{lineno}
\usepackage{algpseudocode}
\usepackage{amsmath,amsfonts,subfigure}
\usepackage{array,url,longtable}
\usepackage{booktabs}
\usepackage{bbding}
\usepackage{multirow}
\usepackage{color, soul}
\sethlcolor{yellow}
\usepackage{subfiles}

\title{Exploring Cross-Domain Few-Shot Classification via Frequency-Aware Prompting}

\author{
Tiange Zhang\and
Qing Cai$^{*}$\and
Feng Gao\and
Lin Qi\And
Junyu Dong\thanks{Corresponding authors: Qing Cai, Junyu Dong.}\\
\affiliations
Faculty of Computer Science and Technology, Ocean University of China\\
\emails
zhangtiange@stu.ouc.edu.cn,
\{cq, gaofeng, qilin, dongjunyu\}@ouc.edu.cn
}

\begin{document}

\maketitle

\begin{abstract}
Cross-Domain Few-Shot Learning has witnessed great stride with the development of meta-learning. However, most existing methods pay more attention to learning domain-adaptive inductive bias (meta-knowledge) through feature-wise manipulation or task diversity improvement while neglecting the phenomenon that deep networks tend to rely more on high-frequency cues to make the classification decision, which thus degenerates the robustness of learned inductive bias since high-frequency information is vulnerable and easy to be disturbed by noisy information. Hence in this paper, we make one of the first attempts to propose a Frequency-Aware Prompting method with mutual attention for Cross-Domain Few-Shot classification, which can let networks simulate the human visual perception of selecting different frequency cues when facing new recognition tasks. Specifically, a frequency-aware prompting mechanism is first proposed, in which high-frequency components of the decomposed source image are switched either with normal distribution sampling or zeroing to get frequency-aware augment samples. Then, a mutual attention module is designed to learn generalizable inductive bias under CD-FSL settings. More importantly, the proposed method is a plug-and-play module that can be directly applied to most off-the-shelf CD-FLS methods. Experimental results on CD-FSL benchmarks demonstrate the effectiveness of our proposed method as well as robustly improve the performance of existing CD-FLS methods. Resources at \url{https://github.com/tinkez/FAP_CDFSC}.
\end{abstract}

\section{Introduction}
Recently, most progress on Few-Shot Learning (FSL) has followed a Meta-Learning pipeline, where the sampled classes of episodic source and target tasks are disjoint but from the same dataset, i.e., they share a similar domain. However, this underlying assumption is not applicable to real-world scenarios since the source and target domains are usually different in realistic tasks. Besides, recent works~\cite{chen2019closerfewshot,guo2020broader} have also revealed that most existing meta-learning models suffer performance degradation when facing a domain shift between source and target tasks, and even underperform simple pre-training and fine-tuning paradigms. This intractable issue makes the new research direction, Cross-Domain Few-Shot Learning (CD-FSL) distinguishable from the conventional FSL settings.

\begin{figure}[t]
  \centering
  \includegraphics[width=0.70\columnwidth]{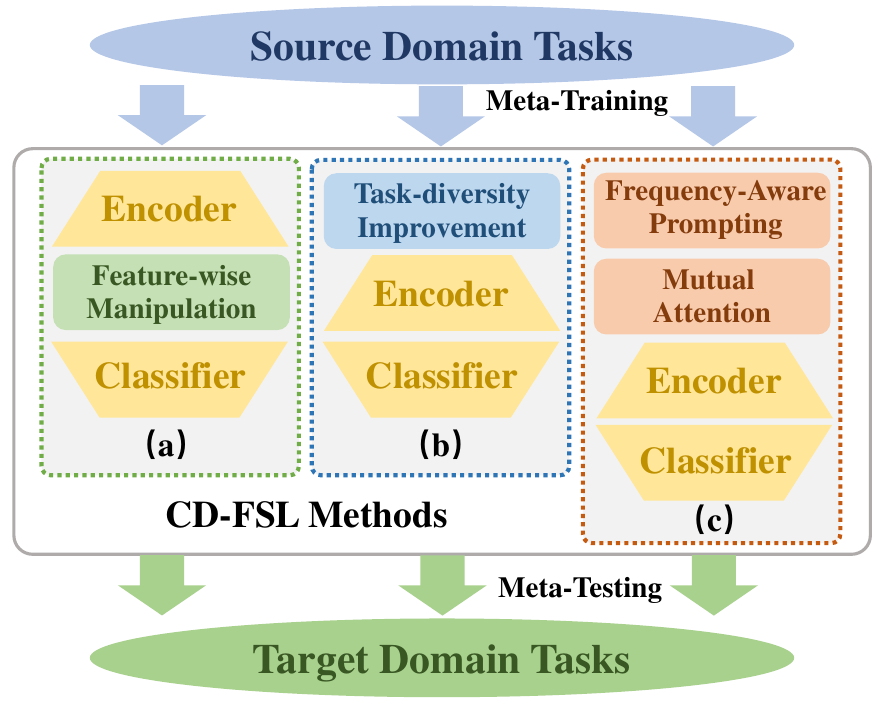}
  \caption{Comparison of existing CD-FSL methods with ours. (a) Feature-wise Manipulation methods have changed the backbone while (b) Task-diversity Improvement methods as well as (c) ours haven't. Thus, (c) is compatible with (a) and (b).}
  \label{IntroCompare}
\end{figure}

CD-FSL aims to generalize FSL models across different domains. Here we focus on the most challenging CD-FSL setting where only one single source domain is available for training. To achieve that,~\cite{tseng2020FWT} uses the feature-wise transformation layers to manipulate the feature activation with affine parameters to improve the generalization of metric-based inductive bias. This feature-wise transformation approach has influenced some later research on varying the feature distribution for target domain adaptation~\cite{liang2021boosting,zhang2022free,hu2022AFA}, but they either have employed extra parameters of weights on backbone networks or suffered from an incompatibility with other models. Apart from the feature-wise manipulation, it is also intuitive that improving the diversity of training tasks also takes a significant role in cross-domain generalization for inductive bias since it's learned from a collection of training tasks. Therefore, ~\cite{wang2021ATA} first resorted to solving a worst-case problem around the source task distribution and proposed Adversarial Task Augmentation (ATA) to construct `challenging' virtual tasks as a task augmentation technique. Their strategy improves the robustness of various inductive biases learned from meta-learning models and achieves good performance on the CD-FSL benchmarks.

However, when tackling the generalization problem on domain shift, most previous CD-FSL methods neglect an important factor that the frequency components of images also get involved in the robustness of inductive bias learned by existing meta-learning models. Several researchers~\cite{wang2020HFC,geirhos2018ImagenetTrainedCNNs,li2021shapetexture,rahaman2019spectral} have already shown that the generalization behaviors of deep neural networks (especially CNNs) are sometimes counter-intuitive to human understandings (e.g., adversarial examples~\cite{nguyen2015deep}). One convincing explanation is that deep neural networks are more sensitive to high-frequency components of a natural image which are not easy-perceivable for human beings in recognition tasks~\cite{wang2020HFC,rahaman2019spectral}. And the experimental observation in~\cite{wang2020HFC} also demonstrates a phenomenon that when making the final classification decision, deep-learning models tend to capture these specific high-frequency cues of a specific training domain for a higher accuracy after they learn from the low-frequency information at the early training stage. This phenomenon will guide the meta-learning models to learn a distribution-specific inductive bias towards the high-frequency cues during the meta-training stage on the source domain, and thus limit the performance of cross-domain generalization when meta-testing on a new domain dataset.

Therefore, in order to learn the more robust inductive bias that can be generalized to the cross-domain scenario for existing meta-learning methods, we make one of the first attempts to consider the CD-FSL challenge from a frequency-aware perspective. As shown in Figure~\ref{IntroCompare}, in this paper, we try to propose a frequency-aware prompting method with mutual attention to prompt the network either to stress the low-frequency semantic information or weaken the tendency on capturing the distribution-specific high-frequency cues, which aims to learn robust inductive bias and let the meta-learning models perceive more like the human when facing a new recognition task. Concretely, we first decompose the source images into corresponding low-frequency and high-frequency components by applying the Discrete Wavelet Transform (DWT) and constructing the frequency-aware augment samples by modifying high-frequency components. In addition, a mutual attention module is designed to enhance the interaction between different frequency-constructed features, and implicitly stress the low-frequency semantic information or suppress the high-frequency cues. Note that~\cite{fu2022waveSAN} also explores the CD-FSL problem from the frequency viewpoint, but their research focuses on varying the feature distribution on a feature-wise level, whereas our frequency-aware prompting method is used at the task level and can be orthogonal with the most off-the-shelf CD-FSL method. We also employ an adversarial training procedure to make task augmentation, which is similar to~\cite{wang2021ATA}. A KL Divergency loss is also applied to restrict the final class prediction of different frequency-reconstructed inputs from the same source image and help the network converge.

Overall, our contribution can be summarized as followed:
\begin{itemize}
\item We consider the most challenging single source domain generalization of the Cross-Domain Few-Shot classification from a frequency-aware perspective. To learn more robust domain-adaptive inductive bias for existing meta-learning models, we proposed a frequency-aware prompting method to simulate human perception when facing a new recognition task. The augment samples are elaborately constructed using the decomposed frequency components to prompt the meta-training process.  
\item A special mutual attention module is designed to facilitate the information interaction across different frequency-reconstructed features and thus promote the network to implicitly stress the related low-frequency semantic information which helps with generalization.  
\item Experimental results on two CD-FSL benchmarks~\cite{guo2020broader,tseng2020FWT} reveal that our method can effectively improve the average classification accuracy in CD-FSL settings when compared to previous methods, and also can be used as a plug-and-play method for most existing meta-learning models.
\end{itemize}

\begin{figure*}[htbp]
  \centering
  \includegraphics[width=0.97\linewidth]{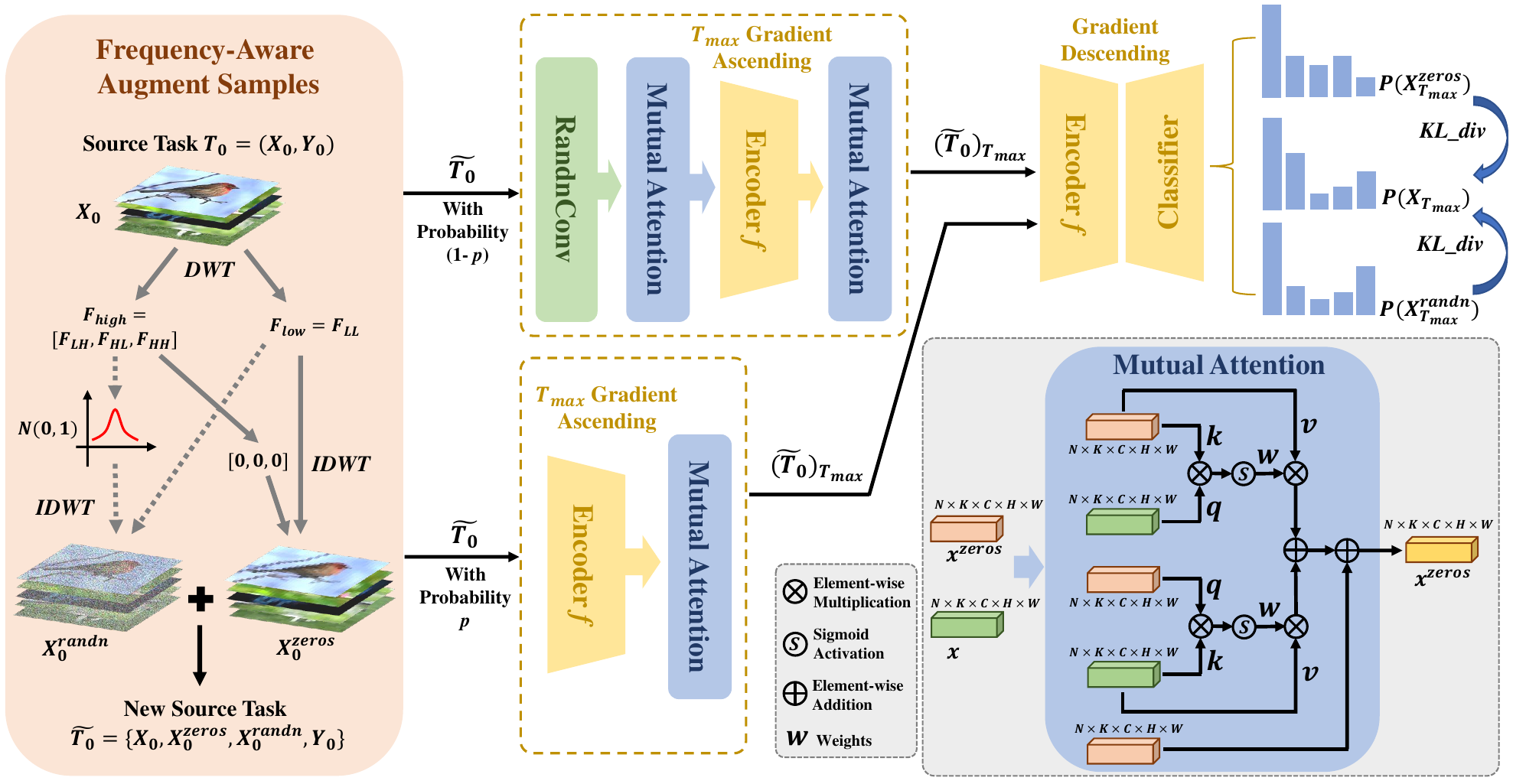}
  \caption{The overall pipeline of the proposed Frequency-Aware Prompting method. $DWT$ represents the Discrete Wavelet Transform, $IDWT$ represents the Inverse Discrete Wavelet Transform, $N(0,1)$ represents the Normal Distribution Sampling, $[0,0,0]$ represents zeroing, $P(.)$ represents the network prediction confidences, and $KL\underline{~} div$ represents the KL divergence loss. An illustration of the Mutual Attention Module is also provided. Detailed explanations of our proposed Frequency-Aware Augment Samples and Mutual Attention are described in subsections 3.3 and 3.4 respectively.}
  \label{Overall Pipeline}
\end{figure*}

\section{Related Works}
In this section, we briefly introduce the previous related works on Few-Shot Learning (FSL) and extended Cross-Domain Few-Shot Learning (CD-FSL).

\subsection{Few-Shot Learning}
Few-Shot Learning~~\cite{shu2018small,wang2020generalizing,lu2020learning,koch2015siamese} has been widely studied, especially in the Meta-Learning manner~\cite{vilalta2002perspective,hospedales2021meta,Sun2019MetaTransferLearning,triantafillou2019MetaDataset,vanschoren2018meta}. Most recent works of meta-learning methods proposed for FSL problems can be roughly divided into two types: 1) Optimization-based methods~\cite{finn2017MAML,nichol2018onfirstorder,ravi2017optimizationAsAModel} aim to learn a generalizable model initialization that can be quickly adapted to target novel tasks with limited labeled samples. 2) Metric-based methods~\cite{vinyals2016MatchingNet,sung2018RelationNet,snell2017prototypical,garcia2018fewshotGNN,oreshkin2018tadam} try to construct a metric function in the embedding space and perform classification on a novel task by comparing the similarity between query images and limited labeled support images through this metric. Among these two types, the metric-based methods have drawn more attention from researchers when considering the performance and simplicity. However, what is common to all of these FSL solutions is that they assume base and novel classes are from the same domain. Due to this limitation, the Cross-Domain Few-Shot Learning (CD-FSL) benchmark is proposed to focus on the domain shift problem.

\subsection{Cross-Domain Few-Shot Learning}
As mentioned, CD-FSL~\cite{guo2020broader,tseng2020FWT,du2022hierarchical,li2022crossTaskSpecific,oh2022understandingCDFSL,yuan2022tasklevelssl} has been trying to address a more challenging and practical scenario where the source (base classes) and target (novel classes) image domains are dissimilar, as well as their labels are disjoint. This problem was preliminarily discussed in~\cite{chen2019closerfewshot} when conducting an analysis of existing meta-learning methods in the cross-domain setting.~\cite{guo2020broader} then extend the study to a Broader Study of Cross-Domain Few-Shot Learning (BSCD-FSL) benchmark which consists of four image datasets, including agriculture (CropDisease), satellite (EuroSAT), dermatology (ISIC), and radiological (ChestX) images, along with a decreasing domain similarity to ImageNet.~\cite{tseng2020FWT} also proposed a benchmark with four datasets and use the feature-wise transformation layer to simulate various feature distributions during the training stage and thus encourage existing metric-based meta-learning models to generalize to these unseen domains. Similarly,~\cite{liang2021boosting} then introduces a noise-enhanced supervised autoencoder (NSAE) which can teach the model to capture broader variations of feature distributions and boost the cross-domain generalize capability.

Different from focusing on diversifying feature distributions,~\cite{wang2021ATA} develops an Adversarial Task Augmentation (ATA) algorithm to construct inductive bias adaptive challenging tasks by solving a worst-case problem. This task augmentation method tries to improve the generalization capability of existing meta-learning models from the perspective that how well the inductive bias (a.k.a meta-knowledge) is learned depends on the training task diversity. It is intuitive that a wider task distribution space would cover the domain gap between source and target domains.         

There are also other researches~\cite{phoo2021selftraining,fu2021Metafdmixup,das2022confess,fu2022generalizedMetaFDMixup} that aim to adapt to the novel target domain by leveraging additional unlabelled data~\cite{phoo2021selftraining} or a few labeled data~\cite{fu2021Metafdmixup,das2022confess}, whereas in this paper we consider that there is no access to any target data.

\section{Methodology}

\subsection{Problem Formulation}
For every meta-learning task $T$ consisting of a support set $T_{s}$ and a query set $T_{q}$, it is generally called a $N$-way $K$-shot situation when $T_{s}$ contains $N$ classes with $K$ labeled samples in each class. And $T_{q}$ consists of $Q$ query samples sharing the same classes as in $T_{s}$. Then it can be defined as $T = (T_{s}, T_{q})$, where $T_{s} = {(x_{i}^{s}, y_{i}^{s})}_{i=1}^{N \times K}$ and $T_{q} = {(x_{j}^{q}, y_{j}^{q})}_{j=1}^{Q}$  are sampled from a task distribution $\mathcal {T}_{0}$. As for cross-domain few-shot classification, due to the domain shift scenario, the meta-training dataset (e.g. mini-ImageNet~\cite{vinyals2016MatchingNet}) is different from the meta-testing dataset (e.g. CUB~\cite{welinder2010caltechCUB}). Here we only consider the setting where only one source domain dataset (mini-Imagenet) is used for training while testing on several other different datasets. But notice that every combination of support set $T_{s}$ and a query set $T_{q}$ for a meta-learning task is from the same domain. 
   
\subsection{Overall Mechanism}
Figure~\ref{Overall Pipeline} shows the overall pipeline of our proposed Frequency-Aware Prompting method. Here we first give an overall explanation of our Frequency-Aware Prompting Mechanism and the training strategy. Our method takes a dual-branch framework, where the decision is made by a hyper-parameter probability $p$. A method for constructing frequency-aware augment samples is designed at the task level. The difference between the two branches is that a random convolution~\cite{Lee2020RandomConv} layer is maintained from~\cite{wang2021ATA} for feature embedding in one branch, and followed by the mutual attention module before the next step. The max iteration number $T_{max}$ of gradient ascending serves for the adversarial training. \textbf{Other details can be viewed at the Algorithm. 1 in supplements.} 

\subsection{Frequency-Aware Augmentation}
To prompt the meta-training with frequency information, we design a method of generating frequency-aware augment samples by reconstructing the high-frequency components. This manner includes two meanings. On the one hand, the extra samples will be reunited along with the original samples to form a new source task, thus improving the task diversity for meta-training. On the other hand, these prompt samples aim to weaken the tendency of networks that focus too much on high-frequency cues while paying less attention to low-frequency information. 

To be specific, for a source task $T_{0}=(X_{0}, Y_{0})$, we conduct Discrete Wavelet Transformation (DWT) on $X_{0}$ with the 2D Haar wavelets, and get four frequency subbands denote as $F_{LL}$, $F_{LH}$, $F_{HL}$, $F_{HH}$. We consider the low-frequency components as $F_{low} = F_{LL}$ and keep it unchanged while performing two different operations on the high-frequency components $F_{high} = [F_{LH}, F_{HL}, F_{HH} ]$ as (1) replacing the $F_{high}$ with the same size tensor of zeros; (2) replacing the $F_{high}$ with the same size of tensor consisting of random numbers sampled from the normal distribution. For the first operation, we can get reconstructed $X_{0}^{zeros}$ after applying Inverse Discrete Wavelet Transformation (IDWT) and $X_{0}^{randn}$ in the same way for the second operation. By doing so, we get a new source task $\widetilde{T_{0}} = \lbrace X_{0}, X_{0}^{zeros}, X_{0}^{randn}, Y_{0} \rbrace$ as mentioned before, where the labels $Y_{0}$ are same. 

Compared to the original $T_{0}$, this new $\widetilde{T_{0}}$ is more diverse as a meta-training source task. Besides, the task distribution $\widetilde{ \mathcal{T}_{0} }$ it represents is more faithfully than $\mathcal{T}_{0}$ since it consists of `noisy' samples near the distribution boundary. Note that for images from the same category, these frequency-aware samples will not introduce semantic drift since we keep the low-frequency components unchanged. This can implicitly prompt the network to rely more on low-frequency information to make the classification decision and learned robust inductive bias which can perform better in the later cross-domain generalization than previous.

By using the gradient ascent process with early stopping (for $T_{max}$ iterations) as the adversarial training based on: 
\begin{equation}
\widetilde{T_{0}} = \lbrace X_{0}, X_{0}^{zeros}, X_{0}^{randn}, Y_{0} \rbrace \,, 
\end{equation}
which is sampled from the wider source task distribution $\widetilde{ \mathcal{T}_{0} }$, we can get corresponding virtual `challenging' tasks as:
\begin{equation}
(\widetilde{T}_{0})_{T_{max}} = \lbrace X_{T_{max}}, X_{T_{max}}^{zeros}, X_{T_{max}}^{randn}, Y_{0} \rbrace \,, 
\end{equation}
where $\widetilde{T}$ can represent a virtual `challenging' task sampled from a distribution $\widetilde{ \mathcal{T} }$ which is $\rho$ distance away from $\widetilde{ \mathcal{T}_{0} }$. 

\subsection{Mutual Attention Module}
To facilitate the information interaction across different frequency-reconstructed features, we proposed a Mutual Attention Module for either stressing the correlated features or suppressing other unrelated information on extra generated features of $x^{zeros}$ and $x^{randn}$ with their original image feature $x$. As shown in the Figure~\ref{Overall Pipeline}, we take $x^{zeros}$ and $x$ as examples for better illustration. The attention module takes two same-size features as inputs, where for every attention operation there is one input used as the query and the other is the key tensor. The attention calculation is mutual to both inputs and the weights are applied to the corresponding value tensor before they are merged. A residual tensor is also added to modify the final outputs.

Figure.~\ref{Overall Pipeline} has shown where to use this attention module. Overall, we consider the features of an original image $x \in \Bbb{R}^{N \times K \times C \times H \times W}$ as an anchor, meanwhile modifying two frequency-manipulated feature $x^{zeros} \in \Bbb{R}^{N \times K \times C \times H \times W}$ and $x^{randn} \in \Bbb{R}^{N \times K \times C \times H \times W}$ based on $x$. The underlying motivation is that the low-frequency features have been preserved and stressed, whereas other uncorrelated frequency features are suppressed. Therefore, the network is implicitly prompted to pay more attention to low-frequency information which is more useful for generalization across different domains. Note that the proposed Mutual Attention Modules do not change the dimension of both inputs and outputs.

\subsection{Training and Inference}

\textbf{Network Training: }Our Frequency-Aware Prompting method is a bi-level optimization process in the meta-training stage, which is similar to~\cite{wang2021ATA}. For the inner loop, it tries to find a `challenging' input based on the new source task for the outer loop by maximizing the loss function, while the outer loop updates the model parameters (learned inductive bias) to minimize the loss function on the new challenging input. And the proposed Frequency-Aware Prompting Augmentation is employed to help improve the diversity of tasks in the inner loop stage. During the meta-training stage, the objective function is first composed of three initial Cross-Entropy (CE) FSL losses from $X_{T_{max}}$, $X_{T_{max}}^{zeros}$ and $X_{T_{max}}^{randn}$ that:  
\begin{equation}
\begin{split}
&\mathcal{L}_{0} = L((X_{T_{max}}, Y_{0});\theta) = CE(X_{T_{max}}, Y_{0}) \,, \\
&\mathcal{L}_{zeros} = L((X_{T_{max}}^{zeros}, Y_{0});\theta) = CE(X_{T_{max}}^{zeros}, Y_{0}) \,, \\
&\mathcal{L}_{randn} = L((X_{T_{max}}^{randn}, Y_{0});\theta) = CE(X_{T_{max}}^{randn}, Y_{0}) \,, \\
\end{split}
\end{equation}
and then to restrict the network prediction and help the network to converge, we also proposed to use the KL divergence loss as the special regularization in training. Specifically, there exist two KL divergence losses defined as:
\begin{equation}
\begin{split}
&\mathcal{L}_{zeros}^{kl} = KLDivergence( P(X_{T_{max}}^{zeros}), P(X_{T_{max}}) ) \,, \\
&\mathcal{L}_{randn}^{kl} = KLDivergence( P(X_{T_{max}}^{randn}), P(X_{T_{max}}) ) \,, \\
\end{split}
\end{equation}
where $P(.)$ represents the distribution of the network prediction scores for classification.

Overall, the training loss can be summarized as: 
\begin{equation}
\begin{split}
\mathcal{L}_{train} = \mathcal{L}_{0} + \mathcal{L}_{zeros} + \mathcal{L}_{randn} + \mathcal{L}_{zeros}^{kl} + \mathcal{L}_{randn}^{kl} \,.
\end{split}
\end{equation}

\noindent\textbf{Network Testing: }During the meta-testing stage of the network, for each episodic task of images sampled from the target domain, we only feed the original images as inputs to the baseline meta-learning backbone with all additional processes discarded, and get their prediction. No frequency-aware augment samples as well as the mutual attention module are used. The class with the highest probability will be considered as the classification results.

\noindent\textbf{Further Explanation of Methodology: }Due to the limited pages, a more detailed explanation of the Methodology can be found in our supplementary materials.

\section{Experiments}
In this section, we evaluate the proposed Frequency-Aware Prompting method on several CD-FSL benchmark datasets and baseline methods. Following we first describe the experiments settings and the implementation details. Then we present the evaluation results and ablation studies.

\subsection{Experiments Settings}
\textbf{Datasets: } We conduct extensive experiments under the very strict cross-domain few-shot learning settings where only one single source dataset is used for meta-training, i.e., mini-ImagneNet~\cite{vinyals2016MatchingNet}. There are two CD-FSL benchmarks used in the meta-testing phase for evaluation. One is from the FWT benchmark~\cite{tseng2020FWT} which includes CUB~\cite{welinder2010caltechCUB}, Cars~\cite{krause20133dCars}, Places~\cite{zhou2017places}, and Plantae~\cite{van2018inaturalistPlantae}. The other one is the BSCD-FSL benchmark~\cite{guo2020broader} which consists of ChestX, ISIC, EuroSAT, and CropDisease. For all experiments, we select the model with the best validation accuracy on mini-ImageNet for the next testing on eight target datasets.

\noindent\textbf{Implementation Details: } We conduct our experiments under the open-source CD-FSL framework provided by~\cite{wang2021ATA}. For a fair comparison, ResNet-10 is selected as the feature extractor and the optimizer is Adam with a fixed learning rate $\alpha = 0.001$. The iteration number for early stopping is kept as $T_{max} = 5$ in all experiments and the gradient ascending step takes a learning rate $\beta$ from $\lbrace 20, 40, 60, 80 \rbrace$. We maintain the Random Convolution layer applied in~\cite{wang2021ATA} with a probability $(1-p)$, where $p$ is selected from $\lbrace 0.5,0.6,0.7 \rbrace$. The filter size $k$ is randomly sampled from the candidate pool $\mathcal{K} = \lbrace 1,3,5,7,11,15 \rbrace$ and the Xavier normal distribution~\cite{glorot2010understandingXavier} is used to initialize the layer weights. Both stride and padding sizes are determined to keep the image size unchanged. Note that we follow the strategy of ATA which uses a pre-trained feature extractor by minimizing the standard cross-entropy classification loss on the 64 training classes in the mini-ImageNet dataset. Evaluation results are obtained in the 5-way 1-shot/5-shot settings with 2000 randomly sampled episodic tasks with 16 query samples per class, and report the average accuracy (\%) with a 95\% confidence interval. All experiments reported are conducted on the RTX 2080Ti GPUs.     

\subsection{Evaluation Results}
Here in this subsection, we present the evaluation results with CD-FSL benchmarks and the baselines.

\noindent\textbf{Comparison with State-of-the-arts: } We follow the settings in~\cite{wang2021ATA} to make a fair comparison on eight datasets with several stat-of-the-arts CD-FSL methods. Specifically, we integrate the Frequency-Aware Prompting (FAP) into three meta-learning baselines: GNN~\cite{garcia2018fewshotGNN}, RelationNet~\cite{sung2018RelationNet}, and TPN~\cite{ma2020transductiveTPN}. Then compare the performance improvements on the baselines using ours with FWT~\cite{tseng2020FWT}, LRP~\cite{sun2021explanationLRP}, and ATA~\cite{wang2021ATA}. Note that LRP can only be applied to metric-based meta-leaning models such as GNN and RelationNet. All the methods presented select ResNet-10 as the feature extractor. Quantitative results of 1-shot and 5-shot classification average accuracies as well as their 95\% confidence interval are shown in Table~\ref{QuantitativeResults}, and the parameters of ours for comparison are shown in Table~\ref{ModelParameters}. As presented, our method achieves considerable accuracy gaining on three baselines, and in most cases achieves the best or second-best performance over benchmark datasets. Also, we observe that our method can provide the most impressive results on the GNN baseline, which means that GNN is more robust when applied with our frequency-aware method under the CD-FSL settings. Furthermore, there are also some performance limitations on the target dataset such as the satellite images (EuroSAT).

\begin{table}[htb]\small
\centering
\begin{tabular}{ccccc}
\hline
baseline    & setting & lr $\beta$ & $T_{max}$ & $p$ \\ \hline
GNN         & 1-shot  & 40  & 5  & 0.5  \\
GNN         & 5-shot  & 40  & 5  & 0.5  \\ 
RelationNet & 1-shot  & 80  & 5  & 0.5  \\
RelationNet & 5-shot  & 80  & 5  & 0.5  \\ 
TPN         & 1-shot  & 20  & 5  & 0.5  \\
TPN         & 5-shot  & 40  & 5  & 0.5  \\ \hline
\end{tabular}
\caption{The parameters of our method in the Table~\ref{QuantitativeResults} presentation.}
\label{ModelParameters}
\end{table}

\begin{table*}[t]\small
\centering
\begin{tabular*}{\linewidth}{l|cccccccc}
\hline
\textbf{5-shot} & ChestX & ISIC  & EuroSAT & CropDisease & Cars  & CUB   & Places & Plantae \\ \hline
GNN (\it{baseline})  & 23.87$\pm$0.2  & 42.54$\pm$0.4  & 78.69$\pm$0.4  & 83.12$\pm$0.4  & 43.70$\pm$0.4  & 62.87$\pm$0.5  & 70.91$\pm$0.5  & 48.51$\pm$0.4 \\
+FWT   & 24.28$\pm$0.2  & 40.87$\pm$0.4  & 78.02$\pm$0.4  & 87.07$\pm$0.4  & 46.19$\pm$0.4  & 64.97$\pm$0.5  & 70.70$\pm$0.5  & 49.66$\pm$0.4 \\
+LRP  & \textbf{24.53$\pm$0.3}  & 44.14$\pm$0.4  & 77.14$\pm$0.4  & 86.15$\pm$0.4  & 46.07$\pm$0.4  & 62.86$\pm$0.5  & 71.38$\pm$0.5  & 50.31$\pm$0.4 \\
+ATA  & 24.32$\pm$0.4  & \textbf{44.91$\pm$0.4}  & \textbf{83.75$\pm$0.4}  & \textbf{90.59$\pm$0.3}  & \textbf{49.14$\pm$0.4}  & \textbf{66.22$\pm$0.5}  & \textbf{75.48$\pm$0.4}  & \textbf{52.69$\pm$0.4} \\
+FAP (\it{ours}) & \textbf{25.31$\pm$0.2}  & \textbf{47.60$\pm$0.4}  & \textbf{82.52$\pm$0.4}  & \textbf{91.79$\pm$0.3}  & \textbf{50.20$\pm$0.4}  & \textbf{67.66$\pm$0.5}  & \textbf{74.98$\pm$0.4}  & \textbf{54.54$\pm$0.4} \\ \hline
RelationNet (\it{baseline})  & 24.07$\pm$0.2  & 38.60$\pm$0.3  & 65.56$\pm$0.4  & 72.86$\pm$0.4  & 40.46$\pm$0.4  & 56.77$\pm$0.4  & 64.25$\pm$0.4  & 42.71$\pm$0.3 \\
+FWT   & 23.95$\pm$0.2  & 38.68$\pm$0.3  & 69.13$\pm$0.4  & 75.78$\pm$0.4  & 40.18$\pm$0.4  & \textbf{59.77$\pm$0.4}  & 65.55$\pm$0.4  & 44.29$\pm$0.3  \\
+LRP  & \textbf{24.28$\pm$0.2}  & \textbf{39.97$\pm$0.3}  & 67.54$\pm$0.4  & 74.21$\pm$0.4  & 41.21$\pm$0.4  & 57.70$\pm$0.4  & 65.35$\pm$0.4  & 43.70$\pm$0.3  \\
+ATA  & \textbf{24.43$\pm$0.2}  & \textbf{40.38$\pm$0.3}  & \textbf{71.02$\pm$0.4}  & \textbf{78.20$\pm$0.4}  & \textbf{42.95$\pm$0.4}  & \textbf{59.36$\pm$0.4}  & \textbf{66.90$\pm$0.4}  & \textbf{45.32$\pm$0.3}  \\
+FAP (\it{ours}) & 23.71$\pm$0.2  & 39.14$\pm$0.4  & \textbf{69.66$\pm$0.3}  & \textbf{78.23$\pm$0.4}  & \textbf{42.86$\pm$0.4}  & 57.78$\pm$0.4  & \textbf{65.97$\pm$0.4}  & \textbf{45.19$\pm$0.3} \\ \hline
TPN (\it{baseline})  & 22.17$\pm$0.2  & \textbf{45.66$\pm$0.3}  & 77.22$\pm$0.4  & 81.91$\pm$0.5  & 44.54$\pm$0.4  & 63.52$\pm$0.4  & 71.39$\pm$0.4  & 50.96$\pm$0.4 \\
+FWT   & 21.22$\pm$0.1  & 36.96$\pm$0.4  & 65.69$\pm$0.5  & 70.06$\pm$0.7  & 34.03$\pm$0.4  & 58.18$\pm$0.5  & 66.75$\pm$0.5  & 43.20$\pm$0.5 \\
+ATA  & \textbf{23.60$\pm$0.2}  & \textbf{45.83$\pm$0.3}  & \textbf{79.47$\pm$0.3}  & \textbf{88.15$\pm$0.5}  & \textbf{46.95$\pm$0.4}  & \textbf{65.31$\pm$0.4}  & \textbf{72.12$\pm$0.4}  & \textbf{55.08$\pm$0.4} \\
+FAP (\it{ours}) & \textbf{24.15$\pm$0.2}  & 44.58$\pm$0.3  & \textbf{80.24$\pm$0.3}  & \textbf{88.34$\pm$0.3}  & \textbf{47.38$\pm$0.4}  & \textbf{64.17$\pm$0.4}  & \textbf{72.05$\pm$0.4}  & \textbf{53.58$\pm$0.4} \\ \hline
\hline
\textbf{1-shot} & ChestX & ISIC  & EuroSAT & CropDisease & Cars  & CUB   & Places & Plantae \\ \hline
GNN (\it{baseline})  & 21.94$\pm$0.2  & 30.14$\pm$0.3  & 54.61$\pm$0.5  & 59.19$\pm$0.5  & 31.72$\pm$0.4  & 44.40$\pm$0.5  & 52.42$\pm$0.5  & 33.60$\pm$0.4 \\
+FWT   & 22.00$\pm$0.2  & 30.22$\pm$0.3  & 55.53$\pm$0.5  & 60.74$\pm$0.5  & 32.25$\pm$0.4  & \textbf{45.50$\pm$0.5}  & 53.44$\pm$0.5  & 32.56$\pm$0.4 \\
+LRP  & \textbf{22.11$\pm$0.2}  & 30.94$\pm$0.3  & 54.99$\pm$0.4  & 59.23$\pm$0.4  & 31.46$\pm$0.4  & 43.89$\pm$0.5  & 52.28$\pm$0.5  & 33.20$\pm$0.4 \\
+ATA  & \textbf{22.10$\pm$0.2}  & \textbf{33.21$\pm$0.4}  & \textbf{61.35$\pm$0.5}  & \textbf{67.47$\pm$0.5}  & \textbf{33.61$\pm$0.4}  & 45.00$\pm$0.5  & \textbf{53.57$\pm$0.4}  & \textbf{34.42$\pm$0.4} \\
+FAP (\it{ours}) & \textbf{22.36$\pm$0.2}  & \textbf{35.63$\pm$0.4}  & \textbf{62.96$\pm$0.5}  & \textbf{69.97$\pm$0.5}  & \textbf{34.69$\pm$0.4}  & \textbf{46.91$\pm$0.5}  & \textbf{54.41$\pm$0.5}  & \textbf{36.73$\pm$0.4} \\ \hline
RelationNet (\it{baseline})  & 21.95$\pm$0.2  & 30.53$\pm$0.3  & 49.08$\pm$0.4  & 53.58$\pm$0.4  & 30.09$\pm$0.3  & 41.27$\pm$0.4  & 48.16$\pm$0.5  & 31.23$\pm$0.3 \\
+FWT   & 21.79$\pm$0.2  & 30.38$\pm$0.3  & 53.53$\pm$0.4  & 57.57$\pm$0.5  & 30.45$\pm$0.3  & \textbf{43.33$\pm$0.4}  & \textbf{49.92$\pm$0.5}  & 32.57$\pm$0.3  \\
+LRP  & \textbf{22.11$\pm$0.2}  & \textbf{31.16$\pm$0.3}  & 50.99$\pm$0.4  & 55.01$\pm$0.4  & 30.48$\pm$0.3  & 41.57$\pm$0.4  & 48.47$\pm$0.5  & 32.11$\pm$0.3  \\
+ATA  & \textbf{22.14$\pm$0.2}  & \textbf{31.13$\pm$0.3}  & \textbf{55.69$\pm$0.5}  & \textbf{61.17$\pm$0.5}  & \textbf{31.79$\pm$0.3}  & \textbf{43.02$\pm$0.4}  & \textbf{51.16$\pm$0.5}  & \textbf{33.72$\pm$0.3}  \\
+FAP (\it{ours}) & 21.56$\pm$0.2  & 30.55$\pm$0.3  & \textbf{53.72$\pm$0.4}  & \textbf{57.97$\pm$0.5}  & \textbf{31.10$\pm$0.3}  & 41.76$\pm$0.4  & 49.61$\pm$0.5  & \textbf{32.98$\pm$0.3} \\ \hline
TPN (\it{baseline})  & 21.05$\pm$0.2  & \textbf{35.08$\pm$0.4}  & \textbf{63.90$\pm$0.5}  & 68.39$\pm$0.6  & 32.42$\pm$0.4  & 48.03$\pm$0.4  & 56.17$\pm$0.5  & \textbf{37.40$\pm$0.4} \\
+FWT   & 20.46$\pm$0.1  & 29.62$\pm$0.3  & 52.68$\pm$0.6  & 56.06$\pm$0.7  & 26.50$\pm$0.3  & 44.24$\pm$0.5  & 52.45$\pm$0.5  & 32.46$\pm$0.4 \\
+ATA  & \textbf{21.67$\pm$0.2}  & \textbf{34.70$\pm$0.4}  & \textbf{65.94$\pm$0.5}  & \textbf{77.82$\pm$0.5}  & \textbf{34.18$\pm$0.4}  & \textbf{50.26$\pm$0.5}  & \textbf{57.03$\pm$0.5}  & \textbf{39.83$\pm$0.4} \\
+FAP (\it{ours}) & \textbf{21.56$\pm$0.2}  & 33.63$\pm$0.4  & 62.62$\pm$0.5  & \textbf{76.11$\pm$0.5}  & \textbf{34.39$\pm$0.4}  & \textbf{50.56$\pm$0.5}  & \textbf{57.34$\pm$0.5}  & \textbf{37.44$\pm$0.4} \\ \hline
\end{tabular*}
\caption{Quantitative results with state-of-the-arts on CD-FSL benchmark datasets. GNN, RelationNet, and TPN are three baselines. The results of other methods are borrowed from~\protect\cite{wang2021ATA}. The training parameters of our results on each baseline presented here are demonstrated in the previous subsection for clearness. The \textbf{Bold} for the best and the second-best results.}
\label{QuantitativeResults}
\end{table*}

\begin{table*}[t]\small
\centering
\begin{tabular*}{\linewidth}{c|cccccccc}
\hline
\textbf{5-shot} & ChestX & ISIC  & EuroSAT & CropDisease & Cars  & CUB   & Places & Plantae \\ \hline
GNN + FAP ($T_{n}$) & 25.31$\pm$0.2  & 47.60$\pm$0.4  & 82.52$\pm$0.4  & 91.79$\pm$0.3  & 50.20$\pm$0.4  & 67.66$\pm$0.5  & 74.98$\pm$0.4  & 54.54$\pm$0.4  \\
GNN + FAP ($T_{n}^{zeros}$)  & 25.14$\pm$0.2  & 48.02$\pm$0.4  & 82.32$\pm$0.4  & 91.06$\pm$0.3  & 49.77$\pm$0.4  & 67.67$\pm$0.5  & 74.97$\pm$0.4  & 53.74$\pm$0.4  \\
GNN + FAP ($T_{n}^{randn}$)   & 24.36$\pm$0.2  & 46.84$\pm$0.4  & 81.63$\pm$0.4  & 90.47$\pm$0.3  & 50.27$\pm$0.4  & 65.83$\pm$0.5  & 74.25$\pm$0.4  & 53.54$\pm$0.4  \\
GNN + FAP ($T_{n}^{noise}$)  & 23.18$\pm$0.2  & 42.11$\pm$0.3  & 71.43$\pm$0.4  & 79.84$\pm$0.5  & 47.11$\pm$0.4  & 54.96$\pm$0.5  & 62.00$\pm$0.5  & 48.60$\pm$0.4  \\  \hline
GNN + ATA ($T_{n}$) & 24.38$\pm$0.2  & 46.31$\pm$0.4  & 83.86$\pm$0.4  & 90.16$\pm$0.3  & 48.99$\pm$0.4  & 64.65$\pm$0.5  & 74.47$\pm$0.4  & 52.28$\pm$0.4  \\
GNN + ATA ($T_{n}^{zeros}$) & 24.97$\pm$0.2  & 46.85$\pm$0.4  & 83.33$\pm$0.4  & 90.34$\pm$0.3  & 49.63$\pm$0.4  & 65.10$\pm$0.5  & 73.68$\pm$0.5  & 51.59$\pm$0.4  \\ 
GNN + ATA ($T_{n}^{randn}$)  & 22.82$\pm$0.2  & 38.00$\pm$0.3  & 72.94$\pm$0.4  & 80.60$\pm$0.5  & 45.02$\pm$0.4  & 47.63$\pm$0.4  & 56.82$\pm$0.5  & 43.26$\pm$0.4  \\
GNN + ATA ($T_{n}^{noise}$)   & 22.97$\pm$0.2  & 37.23$\pm$0.3  & 71.42$\pm$0.5  & 75.95$\pm$0.5  & 43.50$\pm$0.4  & 44.46$\pm$0.4  & 54.74$\pm$0.4  & 40.90$\pm$0.4  \\ \hline
\end{tabular*}
\caption{Robustness evaluation results on different reconstructed meta-testing tasks. $T_{n}$ represents the original target domain task, $T_{n}^{zeros}$, $T_{n}^{randn}$, and $T_{n}^{noise}$ are new target domain meta-testing tasks that described in previous for details. Both two methods are meta-trained with the same parameters as gradient ascending learning rate $\beta$ of 40, $T_{max}$ of 5, and the $p$ with 0.5.}
\label{RobustnessEvaluation}
\end{table*}

\noindent\textbf{Evaluation on Robustness and Frequency Perception: } In order to verify the robustness of the inductive bias learned from our Frequency-Aware Prompting method, we also provide the evaluation results of meta-testing with different frequency-reconstructed images in Table~\ref{RobustnessEvaluation}. To be specific, instead of meta-testing on tasks sampled from the eight novel target datasets, we turn to change the frequency cues of original meta-testing samples into new frequency-reconstructed tasks. For a target domain novel task $T_{n} = (X_n, Y_n)$, we design three types of new target domain task as (1) replacing the $F_{high}$ with zeros and get $T_{n}^{zeros} = (X_n^{zeros}, Y_n)$; (2) replacing the $F_{high}$ with random numbers sampled from the normal distribution and get $T_{n}^{randn} = (X_n^{randn}, Y_n)$; (3) directly add the Gaussian noise to images and get the $T_{n}^{noise} = (X_n^{noise}, Y_n)$. The purpose is to test the robustness of the learned inductive bias when facing a new task with frequency-disturbed cues or noise under the cross-domain settings and its tendency on frequency perception compared to the original meta-testing.

From the results, we learn that the low-frequency components take the most cues for semantic information since using $T_{n}^{zeros}$ for meta-testing can already get a considerable classification accuracy in CD-FSL benchmarks compared to using original tasks $T_{n}$. Besides, the disturbance performed on high-frequency components actually influences the final classification results, which leads to the performance degradation. But our learned inductive bias can be more robust to these disturbances, suggesting that our frequency-aware mechanism prompts the model to concentrate on the most important low-frequency cues. And the testing results presented by using $T_{n}^{noise}$ also demonstrate the robustness of inductive bias learned by our method in CD-FSL settings.

\subsection{Ablation Studies}

\noindent\textbf{Effect of Gradient Ascend Learning Rate $\beta$: } In all the above experiments, we keep the max iteration number for early stopping as $T_{max} = 5$ and the probability of dual-branch decision as $p = 0.5$ unchanged. Thus, the most important hyper-parameter which influenced the training process is the learning rate $\beta$. Here in this ablation study, we try to explore the effect of the learning rate $\beta$ used in the gradient ascend process, which aims to construct a virtual `challenging' task $(\widetilde{T_{0}})_{T_{max}}$ based on the frequency-aware augmented task $\widetilde{T_{0}}$ for gradient descending optimization. 

\begin{figure}[htb]
  \centering
  \includegraphics[width=0.76\columnwidth]{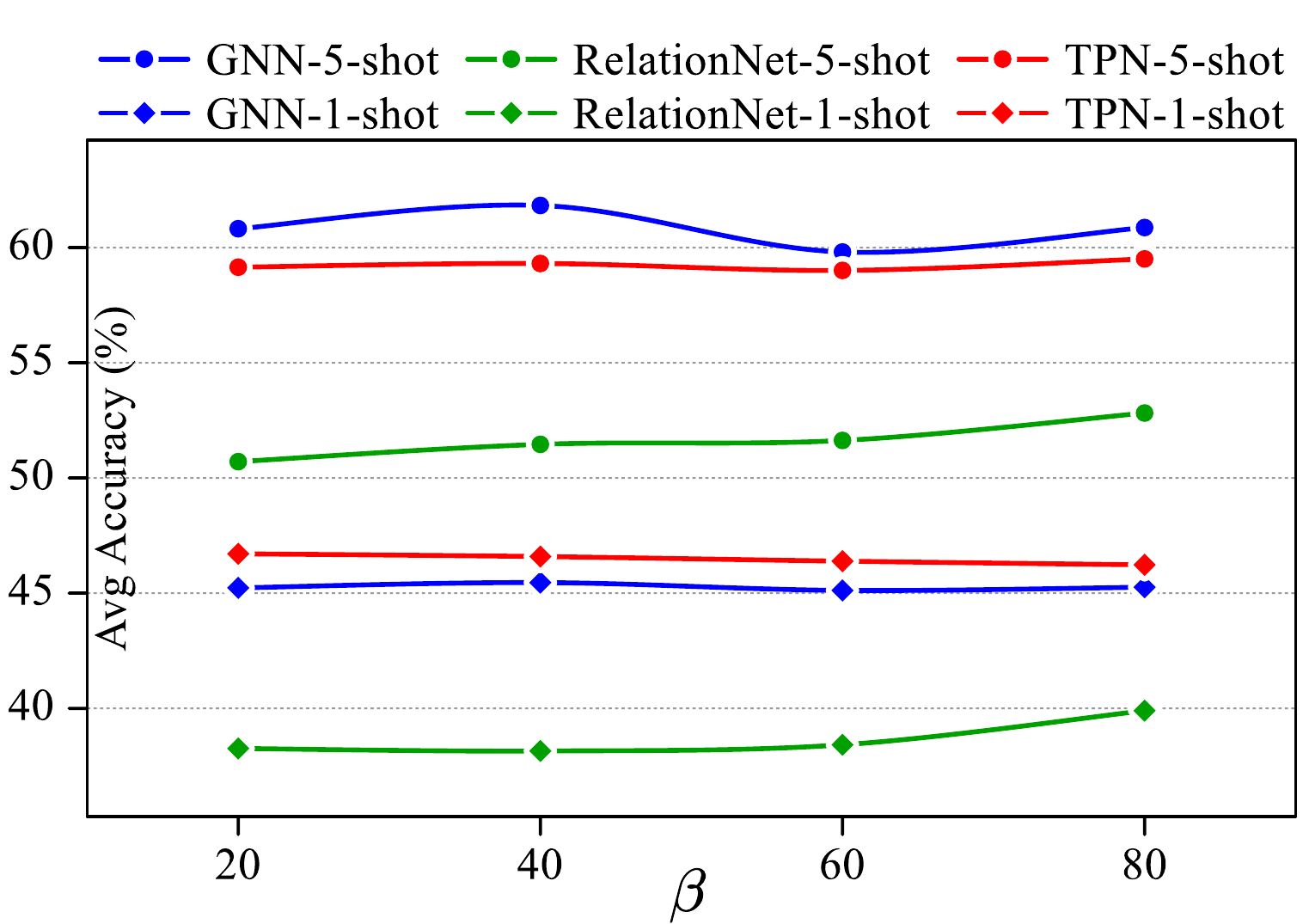}
  \caption{Average accuracy on 5-way 5-shot/1-shot tasks over eight CD-FSL datasets on three baselines with the varying learning rate $\beta$ of gradient ascending process used in the adversarial training. Other parameters are fixed as $T_{max} =5$ and $p = 0.5$ for all stages.}
  \label{Ablation_Maxlr}
\end{figure}

Experiments are conducted on both 5-way 5-shot and 5-way 1-shot settings with the learning rate $\beta$ selected from the set of $\lbrace 20, 40, 60, 80 \rbrace$ as mentioned before. Other meta-training parameters are fixed as $T_{max} =5$ and $p = 0.5$ following the previous experiments. Average classification results over eight cross-domain few-shot learning datasets are reported in Figure~\ref{Ablation_Maxlr}. From it we can see that different baselines prefer various $\beta$. GNN and TPN perform better on a small $\beta$ while the RelationNet needs a larger one.

\noindent\textbf{Effect of Probability $p$: }The average accuracy on eight datasets of our methods is shown in Table~\ref{AvgAccP}. In order to make a fair comparison through different baselines (GNN, RealtionNet, TPN), we set hyper-parameter $p$ as 0.5 unchanged. However, as $p$ grows, the training consumption is increasing according to Algorithm 1 in Supplementary Material. Therefore, keeping $p$ as 0.5 is the better choice to get a performance and effectiveness trade-off in our experiments.

\begin{table}[h]\small
\centering
\begin{tabular}{cccccc}
\hline
$p$   & baseline & lr $\beta$ & $T_{max}$ & 5-shot acc & 1-shot acc \\ \hline
0.5   & GNN & 40 & 5 & 61.82  & 45.46  \\
0.6   & GNN & 40 & 5 & 60.83  & 45.90  \\ 
0.7   & GNN & 40 & 5 & 61.36  & 46.15  \\ \hline
\end{tabular}
\caption{Average accuracy variation over eight CD-FSL datasets with the changing probability $p$.}
\label{AvgAccP}
\end{table}

\begin{figure}[htb]
  \centering
  \includegraphics[width=0.99\columnwidth]{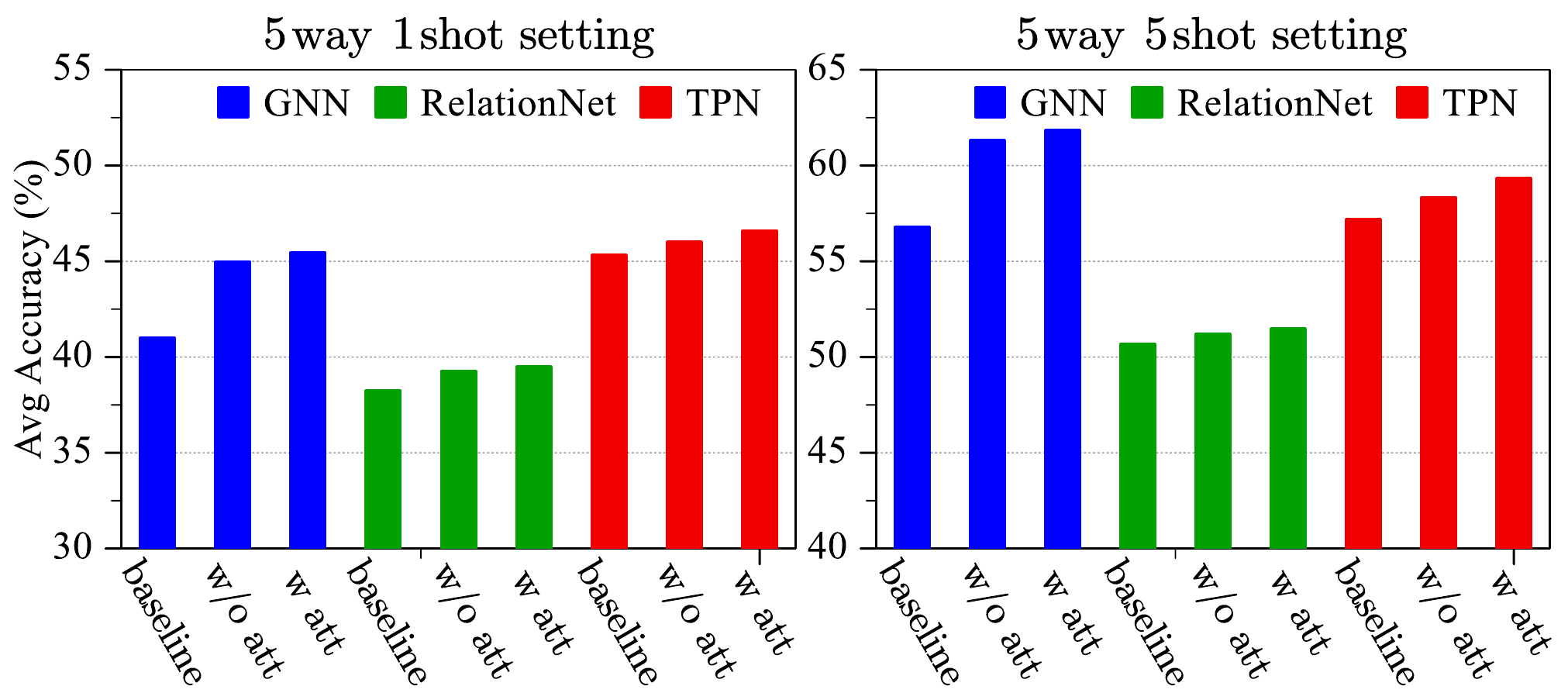}
  \caption{Average accuracy on 5-way 5-shot/1-shot tasks over eight CD-FSL datasets on three baselines. It respectively shows the results of the original baseline, without mutual attention modules (w/o att) and our complete Frequency-Aware Prompting method (w att).}
  \label{Ablation_Attention}
\end{figure}

\noindent\textbf{Effect of Mutual Attention Module: } We also explore the improvement provided by the Mutual Attention Module designed in our method. Ablation experiments are conducted on both 5-way 5-shot and 5-way 1-shot settings with fixed meta-training parameters as $\beta = 40$, $T_{max} =5$, and $p = 0.5$. All baselines and the ablative models are trained in the same manner for a fair comparison. Figure~\ref{Ablation_Attention} shows the average classification results over eight cross-domain few-shot learning datasets with the settings of using baseline model, using frequency-aware prompting method without mutual attention modules, and using our complete method respectively. From the figure we can learn that either with or without the mutual attention modules, the average classification results of the frequency-aware prompting method on eight datasets still achieve higher accuracy than original baseline models. 

\begin{figure}[htb]
  \centering
  \includegraphics[width=0.99\columnwidth]{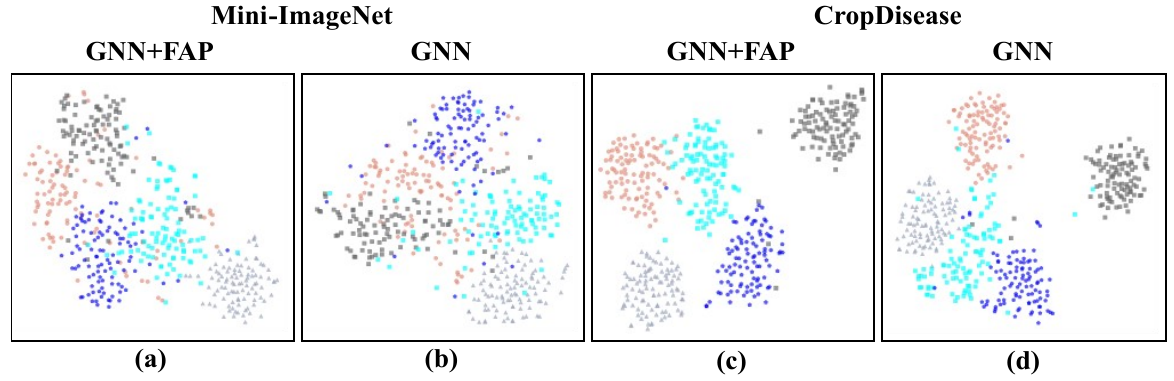}
  \caption{The t-SNE visualization. (a), (b) denote mini-ImageNet while (c), (d) represent CropDisease dataset.}
  \label{TSNE}
\end{figure}


\noindent\textbf{T-SNE Visualization: }Figure~\ref{TSNE} provides the t-SNE visualization of baseline model GNN and GNN+FAP (ours) on both mini-ImageNet (source) and CropDisease (target) datasets. Concretely, the representations encoded by the feature extractor are projected into a 2D space. For each dataset, 5 categories of 100 samples are randomly sampled, and they are fed into two models samely. Different colors indicate different categories. From Figure~\ref{TSNE}, we can learn that compared to GNN, the distribution of each category is indeed more dispersed in our method as expected, hence making it easier to recognize the target categories.

\noindent\textbf{Computational Burden Explanation: }Generally, the proposed method mainly focuses on frequency-aware augmentation when compared with other methods. Therefore, it didn't introduce too much computational burden.

\noindent\textbf{Supplements: } The full paper with supplementary materials can be found in the resources link for further explanation.

\section{Conclusion}
In this paper, we try to explore the cross-domain few-shot learning problem from a frequency-aware perspective. For this purpose, we propose a frequency-aware prompting method with mutual attention to simulating the human perception of selecting frequency cues when facing a new task. Concretely, the generated frequency-aware augment samples can improve the diversity of the meta-training episodic task and prompt the meta-training process during training. The mutual attention modules can facilitate the information interaction across different frequency-reconstructed features to either stress or suppress the corresponding information. Experimental results on the benchmarks show the effectiveness of our method on CD-FSL settings and the robustness of inductive bias learned from existing meta-learning baselines.

\appendix

\section*{Acknowledgments}
This work was supported in part by the National Science and Technology Major Project of China (No.2022ZD0117201); in part by the National Science Foundation of China (No.62102338); in part by TaiShan Scholars Youth Expert Program of Shandong Province (No.tsqn202312109); in part by the China Postdoctoral Science Foundation (No.2023M733342); in part by the Qingdao Postdoctoral Innovation Project (No.QDBSH20230101001). 

\bibliographystyle{named}
\bibliography{ijcai24}

\newpage

\newcommand{\AppendixTitle}[1]{
    \begingroup
    \vspace{1.2cm} 
    \centering
    \LARGE\bfseries #1\par
    \vspace{1.5cm} 
    \endgroup
}

\twocolumn[
\begin{@twocolumnfalse}
\AppendixTitle{Supplementary Material for Exploring Cross-Domain Few-Shot Classification via Frequency-Aware Prompting}
\end{@twocolumnfalse}
]

\input{supp.tex}

\end{document}

%% file: supp.tex
\section{Further Explanation on Motivation}
The correlation between the image frequency components and deep-learning model robustness when facing domain variation has been revealed in several previous works~\cite{wang2020HFC,rahaman2019spectral}, where a convincing explanation is derived from the experiments results that deep-learning models tend to capture the specific high-frequency cues of a specific training domain for a higher accuracy after they learn from the low-frequency information at the early training stage. This is somewhat of a shortcut for models. The high and low-frequency information is considered to be defined in the frequency domains. This phenomenon will lead to a performance degeneracy on cross-domain robustness. Here we illustrated the phenomenon for meta-learning models in the following Table~\ref{FurtherMotivationExplanation}. The three inputs $T_{n}$, $T_{n}^{highs}$ and $T_{n}^{lows}$ are sampled from novel categories of the source training domain (mini-ImageNet). They are unseen samples during the meta-training stage. $T_{n}$ denotes original images, $T_{n}^{highs}$ denotes the images only reconstructed with high-frequency components by IDWT, and $T_{n}^{lows}$ denotes the images only reconstructed with low-frequency components. From the results, we can learn that the baseline meta-learning model GNN achieves the highest test accuracy on novel categories of source domain by only feeding with the high-frequency reconstructed tasks. While our proposed method together with another CD-FSL method cannot achieve such high accuracy with only high-frequency information. These results demonstrate the underlying assumption that meta-learning models tend to take a shortcut training strategy on frequency information.

\begin{table}[h]\small
\centering
\begin{tabular}{ccccc}
\hline
input    & setting &  GNN & GNN+ATA & GNN+FAP \\ \hline
$T_{n}$          & 5-shot  & 80.42$\pm$0.4  & 82.60$\pm$0.4  & 83.11$\pm$0.4  \\
$T_{n}^{highs}$  & 5-shot  & 57.38$\pm$0.5  & 49.90$\pm$0.4  & 46.61$\pm$0.4  \\ 
$T_{n}^{lows}$   & 5-shot  & 79.45$\pm$0.4  & 81.38$\pm$0.4  & 82.22$\pm$0.4  \\ \hline
\end{tabular}
\caption{The results of inputs using different frequency information.}
\label{FurtherMotivationExplanation}
\end{table}

\section{Further Explanation of Methodology}
Here in this subsection, we provide the other detailed information of our proposed Frequency-Aware Prompting method. We first provide the theoretical explanation on adversarial training used for improving the task diversity. Then we provide the Algorithm.~\ref{Algorithm1} which gives an detailed overflow of the meta-training pipeline.

\begin{algorithm}\small
\setlength{\baselineskip}{12pt} 
\caption{Frequency-Aware Prompting Method}
\label{Algorithm1}
\textbf{Input}: Source task distribution $\mathcal{T}_{0}$; initial parameters $\theta_{0}$\\
\textbf{Require}: Gradient descending learning rate $\alpha$ and gradient ascending learning rate $\beta$; max iteration number for early stopping $T_{max}$; probability of dual-branch decision $p \in (0,1)$; candidate pool of filter sizes $\mathcal{K}$\\
\textbf{Output}:learned parameters $\theta$
\begin{algorithmic}[1] 
\State {Initialize $\theta \gets \theta_{0}$}
\While{training}
\State Randomly sample source task $T_{0}=(X_{0},Y_{0})$ from $\mathcal{T}_{0}$. Get $X_{0}^{zeros}$ and $X_{0}^{randn}$ from $X_{0}$. Obtain new augmented source task $\widetilde{T_{0}} = \lbrace X_{0}, X_{0}^{zeros}, X_{0}^{randn}, Y_{0} \rbrace$.
    \If{with probability $(1-p)$}
        \State $X_{0}$ $\gets$ \textit{RandConv}$(X_{0}, \mathcal{K})$
        \State $X_{0}^{zeros}$ $\gets$ \textit{RandConv}$(X_{0}^{zeros}, \mathcal{K})$
        \State $X_{0}^{randn}$ $\gets$ \textit{RandConv}$(X_{0}^{randn}, \mathcal{K})$
        \State $X_{0}^{zeros}$ $\gets$ \textit{MutualAttention} $(X_{0}, X_{0}^{zeros})$
        \State $X_{0}^{randn}$ $\gets$ \textit{MutualAttention} $(X_{0}, X_{0}^{randn})$
        \For{i=1,...,$T_{max}$}
        \State $(X_{0})_{i} = (X_{0})_{i-1} + \beta \cdot \nabla_{X_{0}}L(((X_{0})_{i-1}, Y_{0});\theta)$
        \EndFor
        \State $X_{T_{max}} \gets (X_{0})_{T_{max}}$
        \State $X_{T_{max}}^{zeros}$ $\gets$ \textit{MutualAttention} $(X_{T_{max}}, X_{0}^{zeros})$
        \State $X_{T_{max}}^{randn}$ $\gets$ \textit{MutualAttention} $(X_{T_{max}}, X_{0}^{randn})$
        \State \textbf{return} $X_{T_{max}}, X_{T_{max}}^{zeros}, X_{T_{max}}^{randn}$
    \Else {\quad with probability $p$}
        \For{i=1,...,$T_{max}$}
        \State $(X_{0})_{i} = (X_{0})_{i-1} + \beta \cdot \nabla_{X_{0}}L(((X_{0})_{i-1}, Y_{0});\theta)$
        \State $(X_{0}^{zeros})_{i} = (X_{0}^{zeros})_{i-1} +$
        \State $\quad \quad \quad \quad \quad \beta \cdot \nabla_{X_{0}^{zeros}}L(((X_{0}^{zeros})_{i-1}, Y_{0});\theta)$
        \State $(X_{0}^{randn})_{i} = (X_{0}^{randn})_{i-1} +$
        \State $\quad \quad \quad \quad \quad \beta \cdot \nabla_{X_{0}^{randn}}L(((X_{0}^{randn})_{i-1}, Y_{0});\theta)$
        \EndFor
        \State $X_{T_{max}} \gets (X_{0})_{T_{max}}$
        \State $X_{T_{max}}^{zeros} \gets (X_{0}^{zeros})_{T_{max}}$
        \State $X_{T_{max}}^{randn} \gets (X_{0}^{randn})_{T_{max}}$
        \State $X_{T_{max}}^{zeros}$ $\gets$ \textit{MutualAttention} $(X_{T_{max}}, X_{T_{max}}^{zeros})$
        \State $X_{T_{max}}^{randn}$ $\gets$ \textit{MutualAttention} $(X_{T_{max}}, X_{T_{max}}^{randn})$
        \State \textbf{return} $X_{T_{max}}, X_{T_{max}}^{zeros}, X_{T_{max}}^{randn}$
    \EndIf
    \State $\theta \gets \theta - \alpha \cdot \nabla_{\theta}L( \lbrace X_{T_{max}}, X_{T_{max}}^{zeros}, X_{T_{max}}^{randn}, Y_{0} \rbrace;\theta)$
\EndWhile
\State \textbf{return} $\theta$
\end{algorithmic}
\end{algorithm}

\begin{table*}[htb]\small
\centering
\tabcolsep=0.085cm
\begin{tabular*}{\linewidth}{c|c|ccccccccc}
\hline
\textbf{5-shot} & Task & ChestX & ISIC  & EuroSAT & CropDisease & Cars  & CUB   & Places & Plantae & Average \\ \hline
GNN (\it{baseline})  & $T_{n}$ & 23.87$\pm$0.2  & 42.54$\pm$0.4  & 78.69$\pm$0.4  & 83.12$\pm$0.4  & 43.70$\pm$0.4  & 62.87$\pm$0.5  & 70.91$\pm$0.5  & 48.51$\pm$0.4  & 56.78  \\ \hline
\multirow{4}{*}{GNN + FAP} & $T_{n}$ & 25.31$\pm$0.2  & 47.60$\pm$0.4  & 82.52$\pm$0.4  & 91.79$\pm$0.3  & 50.20$\pm$0.4  & 67.66$\pm$0.5  & 74.98$\pm$0.4  & 54.54$\pm$0.4  & \textbf{61.82}  \\
& $T_{n}^{zeros}$  & 25.14$\pm$0.2  & 48.02$\pm$0.4  & 82.32$\pm$0.4  & 91.06$\pm$0.3  & 49.77$\pm$0.4  & 67.67$\pm$0.5  & 74.97$\pm$0.4  & 53.74$\pm$0.4  & \textbf{61.59}  \\
& $T_{n}^{randn}$   & 24.36$\pm$0.2  & 46.84$\pm$0.4  & 81.63$\pm$0.4  & 90.47$\pm$0.3  & 50.27$\pm$0.4  & 65.83$\pm$0.5  & 74.25$\pm$0.4  & 53.54$\pm$0.4  & 60.90  \\
& $T_{n}^{noise}$  & 23.18$\pm$0.2  & 42.11$\pm$0.3  & 71.43$\pm$0.4  & 79.84$\pm$0.5  & 47.11$\pm$0.4  & 54.96$\pm$0.5  & 62.00$\pm$0.5  & 48.60$\pm$0.4  & 53.65  \\  \hline
RelationNet (\it{baseline}) & $T_{n}$ & 24.07$\pm$0.2  & 38.60$\pm$0.3  & 65.56$\pm$0.4  & 72.86$\pm$0.4  & 40.46$\pm$0.4  & 56.77$\pm$0.4  & 64.25$\pm$0.4  & 42.71$\pm$0.3 & 50.66  \\  \hline
\multirow{4}{*}{RelationNet + FAP} & $T_{n}$ & 23.71$\pm$0.2  & 39.14$\pm$0.4  & 69.66$\pm$0.3  & 78.23$\pm$0.4  & 42.86$\pm$0.4  & 57.78$\pm$0.4  & 65.97$\pm$0.4  & 45.19$\pm$0.3  & \textbf{52.82}  \\
& $T_{n}^{zeros}$  & 23.64$\pm$0.2  & 39.32$\pm$0.3  & 69.24$\pm$0.4  & 77.83$\pm$0.4  & 42.75$\pm$0.4  & 58.06$\pm$0.4  & 65.49$\pm$0.4  & 44.18$\pm$0.4  & \textbf{52.56}  \\
& $T_{n}^{randn}$   & 23.45$\pm$0.2  & 39.23$\pm$0.3  & 61.39$\pm$0.4  & 75.43$\pm$0.4  & 42.26$\pm$0.4  & 56.08$\pm$0.4  & 63.91$\pm$0.4  & 43.10$\pm$0.4  & 50.61  \\
& $T_{n}^{noise}$  & 22.38$\pm$0.2  & 34.19$\pm$0.3  & 44.46$\pm$0.4  & 61.68$\pm$0.4  & 39.47$\pm$0.3  & 44.46$\pm$0.4  & 52.73$\pm$0.4  & 37.55$\pm$0.3  & 42.12  \\  \hline
TPN (\it{baseline})  & $T_{n}$ & 22.17$\pm$0.2  & 45.66$\pm$0.3  & 77.22$\pm$0.4  & 81.91$\pm$0.5  & 44.54$\pm$0.4  & 63.52$\pm$0.4  & 71.39$\pm$0.4  & 50.96$\pm$0.4 & 57.17  \\  \hline
\multirow{4}{*}{TPN + FAP} & $T_{n}$ & 24.15$\pm$0.2  & 44.58$\pm$0.3  & 80.24$\pm$0.3  & 88.34$\pm$0.3  & 47.38$\pm$0.4  & 64.17$\pm$0.4  & 72.05$\pm$0.4  &53.58$\pm$0.4  & \textbf{59.31}  \\
& $T_{n}^{zeros}$ & 24.15$\pm$0.2  & 44.80$\pm$0.3  & 80.11$\pm$0.4  & 88.45$\pm$0.3  & 47.23$\pm$0.4  & 64.07$\pm$0.4  & 71.71$\pm$0.4  & 53.78$\pm$0.4  & \textbf{59.29}  \\ 
& $T_{n}^{randn}$  & 23.30$\pm$0.2  & 44.00$\pm$0.3  & 74.31$\pm$0.4  & 87.83$\pm$0.3  & 47.65$\pm$0.4  & 64.12$\pm$0.4  & 71.08$\pm$0.4  & 52.68$\pm$0.4  & 58.12  \\
& $T_{n}^{noise}$   & 22.12$\pm$0.2  & 40.58$\pm$0.3  & 65.44$\pm$0.3  & 81.65$\pm$0.4  & 47.58$\pm$0.4  & 59.27$\pm$0.4  & 65.67$\pm$0.4  & 50.39$\pm$0.4  & 54.09  \\  \hline

\hline
\textbf{1-shot} & Task & ChestX & ISIC  & EuroSAT & CropDisease & Cars  & CUB   & Places & Plantae & Average \\ \hline
GNN (\it{baseline})  & $T_{n}$ & 21.94$\pm$0.2  & 30.14$\pm$0.3  & 54.61$\pm$0.5  & 59.19$\pm$0.5  & 31.72$\pm$0.4  & 44.40$\pm$0.5  & 52.42$\pm$0.5  & 33.60$\pm$0.4 & 41.00  \\ \hline
\multirow{4}{*}{GNN + FAP} & $T_{n}$ & 22.36$\pm$0.2  & 35.63$\pm$0.4  & 62.96$\pm$0.5  & 69.97$\pm$0.5  & 34.69$\pm$0.4  & 46.91$\pm$0.5  & 54.41$\pm$0.5  & 36.73$\pm$0.4  & \textbf{45.46}  \\
& $T_{n}^{zeros}$  & 22.50$\pm$0.2  & 35.05$\pm$0.4  & 63.19$\pm$0.5  & 69.09$\pm$0.6  & 34.44$\pm$0.4  & 47.89$\pm$0.5  & 54.36$\pm$0.5  & 36.32$\pm$0.4  & \textbf{45.36}  \\
& $T_{n}^{randn}$   & 21.95$\pm$0.2  & 35.74$\pm$0.4  & 56.58$\pm$0.5  & 69.24$\pm$0.6  & 34.84$\pm$0.4  & 45.39$\pm$0.5  & 50.97$\pm$0.4  & 35.59$\pm$0.4  & 43.79  \\
& $T_{n}^{noise}$  & 21.15$\pm$0.2  & 29.86$\pm$0.4  & 54.38$\pm$0.5  & 58.65$\pm$0.6  & 32.09$\pm$0.4  & 37.31$\pm$0.4  & 41.85$\pm$0.5  & 31.86$\pm$0.4  & 38.39  \\  \hline
RelationNet (\it{baseline})  & $T_{n}$ & 21.95$\pm$0.2  & 30.53$\pm$0.3  & 49.08$\pm$0.4  & 53.58$\pm$0.4  & 30.09$\pm$0.3  & 41.27$\pm$0.4  & 48.16$\pm$0.5  & 31.23$\pm$0.3 & 38.23  \\  \hline
\multirow{4}{*}{RelationNet + FAP} & $T_{n}$ & 21.56$\pm$0.2  & 30.55$\pm$0.3  & 53.72$\pm$0.4  & 57.97$\pm$0.5  & 31.10$\pm$0.3  & 41.76$\pm$0.4  & 49.61$\pm$0.5  & 32.98$\pm$0.3  & \textbf{39.91}  \\
& $T_{n}^{zeros}$  & 21.69$\pm$0.2  & 30.41$\pm$0.3  & 54.00$\pm$0.5  & 57.12$\pm$0.5  & 31.56$\pm$0.3  & 41.57$\pm$0.4  & 49.00$\pm$0.5  & 33.22$\pm$0.4  & \textbf{39.82}  \\
& $T_{n}^{randn}$   & 21.49$\pm$0.2  & 30.00$\pm$0.3  & 47.65$\pm$0.4  & 57.43$\pm$0.5  & 30.39$\pm$0.3  & 39.34$\pm$0.4  & 48.53$\pm$0.5  & 32.23$\pm$0.3  & 38.38  \\
& $T_{n}^{noise}$  & 20.94$\pm$0.2  & 29.26$\pm$0.3  & 44.53$\pm$0.4  & 45.60$\pm$0.4  & 28.86$\pm$0.3  & 32.07$\pm$0.3  & 37.92$\pm$0.4  & 28.75$\pm$0.3  & 33.49  \\  \hline
TPN (\it{baseline})  & $T_{n}$ & 21.05$\pm$0.2  & 35.08$\pm$0.4  & 63.90$\pm$0.5  & 68.39$\pm$0.6  & 32.42$\pm$0.4  & 48.03$\pm$0.4  & 56.17$\pm$0.5  & 37.40$\pm$0.4 & 45.31  \\  \hline
\multirow{4}{*}{TPN + FAP} & $T_{n}$ & 21.56$\pm$0.2  & 33.63$\pm$0.4  & 62.62$\pm$0.5  & 76.11$\pm$0.5  & 34.39$\pm$0.4  & 50.56$\pm$0.5  & 57.34$\pm$0.5  & 37.44$\pm$0.4  & \textbf{46.71}  \\
& $T_{n}^{zeros}$ & 21.64$\pm$0.2  & 33.63$\pm$0.3  & 62.89$\pm$0.5  & 75.44$\pm$0.5  & 34.58$\pm$0.4  & 50.73$\pm$0.5  & 57.24$\pm$0.5  & 37.45$\pm$0.4  & \textbf{46.70}  \\ 
& $T_{n}^{randn}$  & 21.38$\pm$0.2  & 34.41$\pm$0.3  & 60.05$\pm$0.5  & 75.43$\pm$0.5  & 34.38$\pm$0.4  & 50.35$\pm$0.5  & 56.38$\pm$0.5  & 37.68$\pm$0.4  & 46.26  \\
& $T_{n}^{noise}$   & 21.04$\pm$0.2  & 31.42$\pm$0.3  & 51.77$\pm$0.5  & 65.95$\pm$0.5  & 34.31$\pm$0.4  & 45.77$\pm$0.5  & 51.46$\pm$0.5  & 36.39$\pm$0.4  & 42.26  \\ \hline
\end{tabular*}

\caption{Meta-testing results on different reconstructed target domain tasks. All methods are tested with the same model as presented in the subsection \noindent\textbf{Evaluation on Robustness and Frequency Perception}.}
\label{MoreResultsDifferentReconstructedTargetDomainTasks}
\end{table*}

\begin{table*}[htb]\small
\centering
\begin{tabular*}{\linewidth}{l|cccccccc}
\hline
\textbf{5-shot} & ChestX & ISIC  & EuroSAT & CropDisease & Cars  & CUB   & Places & Plantae \\ \hline
GNN (\it{baseline})  & 23.87$\pm$0.2  & 42.54$\pm$0.4  & 78.69$\pm$0.4  & 83.12$\pm$0.4  & 43.70$\pm$0.4  & 62.87$\pm$0.5  & 70.91$\pm$0.5  & 48.51$\pm$0.4 \\
GNN+FWT   & 24.28$\pm$0.2  & 40.87$\pm$0.4  & 78.02$\pm$0.4  & 87.07$\pm$0.4  & 46.19$\pm$0.4  & 64.97$\pm$0.5  & 70.70$\pm$0.5  & 49.66$\pm$0.4 \\
GNN+FWT+FAA  & \textbf{25.27$\pm$0.2}  & \textbf{44.95$\pm$0.4}  & \textbf{77.54$\pm$0.4}  & \textbf{89.56$\pm$0.4}  & \textbf{46.49$\pm$0.5}  & \textbf{68.73$\pm$0.5}  & \textbf{75.61$\pm$0.5}  & \textbf{55.16$\pm$0.4} \\
GNN+FAP (\it{ours}) & \textbf{25.31$\pm$0.2}  & \textbf{47.60$\pm$0.4}  & \textbf{82.52$\pm$0.4}  & \textbf{91.79$\pm$0.3}  & \textbf{50.20$\pm$0.4}  & \textbf{67.66$\pm$0.5}  & \textbf{74.98$\pm$0.4}  & \textbf{54.54$\pm$0.4} \\ \hline
TPN (\it{baseline})  & 22.17$\pm$0.2  & 45.66$\pm$0.3  & 77.22$\pm$0.4  & 81.91$\pm$0.5  & 44.54$\pm$0.4  & 63.52$\pm$0.4  & 71.39$\pm$0.4  & 50.96$\pm$0.4 \\
TPN+FWT   & 21.22$\pm$0.1  & 36.96$\pm$0.4  & 65.69$\pm$0.5  & 70.06$\pm$0.7  & 34.03$\pm$0.4  & 58.18$\pm$0.5  & 66.75$\pm$0.5  & 43.20$\pm$0.5 \\
TPN+FWT+FAA   & \textbf{21.44$\pm$0.1}  & \textbf{45.80$\pm$0.3}  & \textbf{72.35$\pm$0.4}  & \textbf{77.64$\pm$0.6}  & \textbf{41.96$\pm$0.4}  & \textbf{61.88$\pm$0.4}  & \textbf{71.72$\pm$0.4}  & \textbf{51.88$\pm$0.4} \\
TPN+FAP (\it{ours}) & \textbf{24.15$\pm$0.2}  & \textbf{44.58$\pm$0.3}  & \textbf{80.24$\pm$0.3}  & \textbf{88.34$\pm$0.3}  & \textbf{47.38$\pm$0.4}  & \textbf{64.17$\pm$0.4}  & \textbf{72.05$\pm$0.4}  & \textbf{53.58$\pm$0.4} \\ \hline
\hline
\textbf{1-shot} & ChestX & ISIC  & EuroSAT & CropDisease & Cars  & CUB   & Places & Plantae \\ \hline
GNN (\it{baseline})  & 21.94$\pm$0.2  & 30.14$\pm$0.3  & 54.61$\pm$0.5  & 59.19$\pm$0.5  & 31.72$\pm$0.4  & 44.40$\pm$0.5  & 52.42$\pm$0.5  & 33.60$\pm$0.4 \\
GNN+FWT   & 22.00$\pm$0.2  & 30.22$\pm$0.3  & 55.53$\pm$0.5  & 60.74$\pm$0.5  & 32.25$\pm$0.4  & 45.50$\pm$0.5  & 53.44$\pm$0.5  & 32.56$\pm$0.4 \\
GNN+FWT+FAA   & \textbf{22.06$\pm$0.2}  & \textbf{32.89$\pm$0.3}  & \textbf{60.71$\pm$0.5}  & \textbf{68.53$\pm$0.6}  & \textbf{33.20$\pm$0.4}  & \textbf{48.98$\pm$0.5}  & \textbf{56.30$\pm$0.6}  & \textbf{37.14$\pm$0.4} \\
GNN+FAP (\it{ours}) & \textbf{22.36$\pm$0.2}  & \textbf{35.63$\pm$0.4}  & \textbf{62.96$\pm$0.5}  & \textbf{69.97$\pm$0.5}  & \textbf{34.69$\pm$0.4}  & \textbf{46.91$\pm$0.5}  & \textbf{54.41$\pm$0.5}  & \textbf{36.73$\pm$0.4} \\ \hline
TPN (\it{baseline})  & 21.05$\pm$0.2  & 35.08$\pm$0.4  & 63.90$\pm$0.5  & 68.39$\pm$0.6  & 32.42$\pm$0.4  & 48.03$\pm$0.4  & 56.17$\pm$0.5  & 37.40$\pm$0.4 \\
TPN+FWT   & 20.46$\pm$0.1  & 29.62$\pm$0.3  & 52.68$\pm$0.6  & 56.06$\pm$0.7  & 26.50$\pm$0.3  & 44.24$\pm$0.5  & 52.45$\pm$0.5  & 32.46$\pm$0.4 \\
TPN+FWT+FAA   & \textbf{20.70$\pm$0.1}  & \textbf{34.74$\pm$0.4}  & \textbf{59.31$\pm$0.5}  & \textbf{62.41$\pm$0.6}  & \textbf{29.45$\pm$0.3}  & \textbf{46.82$\pm$0.5}  & \textbf{56.50$\pm$0.5}  & \textbf{36.23$\pm$0.4} \\
TPN+FAP (\it{ours}) & \textbf{21.56$\pm$0.2}  & \textbf{33.63$\pm$0.4}  & \textbf{62.62$\pm$0.5}  & \textbf{76.11$\pm$0.5}  & \textbf{34.39$\pm$0.4}  & \textbf{50.56$\pm$0.5}  & \textbf{57.34$\pm$0.5}  & \textbf{37.44$\pm$0.4} \\ \hline
\end{tabular*}
\caption{Quantitative results of different combination by using Frequency-Aware Prompting (FAP) on the CD-FSL benchmark datasets. Here the FAA denotes the Frequency-Aware Augmentation which remove the adversarial task augmentation process compared with the complete FAP. Results in \textbf{Bold} represent the methods employing the FAA.}
\label{FAA_results}
\label{MoreResultsFAPvFAA}
\end{table*}

Since the inductive bias is derived from the training tasks in meta-learning, its generalization capability to some extent depends on the diversity of the training tasks, and expanding the source task is an intuitive idea for improving the robustness of the inductive bias across domains. Motivated by this, we firstly try to generate extra frequency-aware samples to augment the original source task $T_{0} = (T_{s}, T_{q})$ into a new source task $\widetilde{T_{0}} = (\widetilde{T_{s}}, \widetilde{T_{q}})$.  

In order to keep the continuity in training, a task $T$ can be also represented as the vector concatenated by the support and query samples along with their labels in the task, i.e.,
\begin{equation}
  T = [x_{1}^{s}, y_{1}^{s}, ... , x_{N \times K}^{s}, y_{N \times K}^{s}, x_{1}^{q}, y_{1}^{q}, ... , x_{Q}^{q}, y_{Q}^{q}] \,,
\end{equation}
where $[.,.]$ denotes the concatenation operation. And similarly, the distribution of tasks can be treated as the joint distribution of samples and labels within the task, i.e.,
\begin{equation}
  \mathcal{T}(T) = P(x_{1}^{s}, y_{1}^{s}, ... , x_{N \times K}^{s}, y_{N \times K}^{s}, x_{1}^{q}, y_{1}^{q}, ... , x_{Q}^{q}, y_{Q}^{q})  \,.
\end{equation}

Here the number of samples in task $T$ is assumed to be fixed, so as the dimension of $T$. Let the set of all samples in a task be $X$ and their corresponding labels be $Y$, i.e.,
\begin{equation}
\begin{split}
  X = [x_{1}^{s}, ... , x_{N \times K}^{s}, x_{1}^{q}, ... , x_{Q}^{q}] \,, \\
  Y = [y_{1}^{s}, ... , y_{N \times K}^{s}, y_{1}^{q}, ... , y_{Q}^{q}] \,,
\end{split}
\end{equation}
and then $T$ can be rewritten as $T = (X, Y)$. Similarly, the source task can be also defined as $T_{0} = (X_{0}, Y_{0})$ for simplify, while the new source task is denoted as: 
\begin{equation}
\begin{aligned}
\widetilde{T_{0}} &= \lbrace (X_{0}, Y_{0}), (X_{0}^{zeros}, Y_{0}), (X_{0}^{randn}, Y_{0}) \rbrace  \,, \\
&= \lbrace X_{0}, X_{0}^{zeros}, X_{0}^{randn}, Y_{0} \rbrace \,,
\end{aligned}
\end{equation}
where $X_{0}^{zeros}$ and $X_{0}^{randn}$ are generated frequency-aware augment samples by the proposed method.

After obtaining $\widetilde{T_{0}}$, with the Encoder $f$, we then also follow an adversarial training strategy proposed in~\cite{wang2021ATA} which tries to construct a virtual `challenging' task based on $\widetilde{T_{0}}$ by solving a worst-case problem around the source task distribution $\mathcal {T}_{0}$ that: 
\begin{equation}
  \underset{\theta \in \Theta} \min \underset{D({\mathcal {T}},{\mathcal {T}_{0}}) \leq \rho} \sup \mathbb{E}_{\mathcal {T}}[L^{meta}(\widetilde{T_{q}}, \psi)], \psi = \mathcal {F} (\widetilde{T_{s}}, \theta)  \,,
\end{equation}
where $\theta \in \Theta$ represents the model parameters, $\widetilde{T_{q}}$, $\widetilde{T_{s}}$ are query and support samples from $\widetilde{{T}_{0}}$, $L^{meta}(\widetilde{T_{q}}, \psi)$ is the loss function which depends on the model's inductive bias, and $D({\mathcal {T}},{\mathcal {T}_{0}})$ is the distance between two task distributions. In~\cite{wang2021ATA}, the authors have pointed out that instead of minimizing the loss function on $\mathcal {T}_{0}$, the solution of the worst-case problem described in Equation (5) also guarantees a good performance on a wider space of task distribution $\mathcal {T}$ which is $\rho$ away from the base task distribution $\mathcal {T}_{0}$. Whereas in our method, the original source task distribution is also expanded by the newly generated frequency-aware samples. They can be considered as the interpolation samples lie in the boundary of the $\mathcal{T}_{0}$. Taking these frequency-aware augment samples into consideration will provide a more faithful representation of the source task distribution as $\widetilde{\mathcal{T}_{0}}$, and thus get an, even more, wider task distribution $\widetilde{\mathcal{T}}$ that is $\rho$ distance away as defined.  

Finally, the detailed derivation in~\cite{wang2021ATA} proved that there exists a unique unknown $\hat{T}$ which can serve as a virtual `challenging' task. For an initialization of task $\widetilde{T_{0}}$ and model parameters $\theta_{0}$, it can be obtained by using a gradient ascent process with early stopping. For the $i$-th iteration, the update is: 
\begin{equation}
  X_{i} = X_{i-1} + \beta \cdot \nabla_{X}L((X_{i-1}, Y_{0});\theta) \,,
\end{equation}
where $X$ represents samples, and the label $Y_{0}$ is unchanged. $L(.)$ denotes cross-entropy loss, and $\beta$ represents the learning rate for the SGD optimizer for the gradient ascent process. After $T_{max}$ iterations, the virtual `challenging' task $(\widetilde{T_{0}})_{T_{max}}$ is obtained along with new model parameters $\theta$. For meta-learning models, the virtual task $(\widetilde{T_{0}})_{T_{max}}$ is harder to optimize than source task $\widetilde{T_{0}}$, therefore the inductive bias of the model learned with it tends to be more robust. That means, optimizing the results in Equation (7) equivalent to adaptively solving a virtual `challenging' task for the currently inductive bias. $(\widetilde{T_{0}})_{T_{max}}$ then finally serve as the inputs to Encoder $f$ again for gradient descending that:
\begin{equation}
  \theta \gets \theta - \alpha \nabla_{\theta} L((\widetilde{T_{0}})_{T_{max}};\theta)  \,,
\end{equation}
where $\alpha$ is the learning rate, and $L(.)$ is the whole loss function we have introduced in the paper.

\section{More Experimental Results}

\subsection{Robustness and Frequency Perception}
In this subsection, we provide more experimental results of meta-testing on different reconstructed target domain tasks. Table~\ref{MoreResultsDifferentReconstructedTargetDomainTasks} is an extension of subsection \noindent\textbf{Evaluation on Robustness and Frequency Perception}.

As mentioned before, $T_{n} = (X_n, Y_n)$ represents the standard meta-testing task, while the others are: 1) $T_{n}^{zeros} = (X_n^{zeros}, Y_n)$ where the $F_{high}$ is replaced with a same size tensor of zeros, 2) $T_{n}^{randn} = (X_n^{randn}, Y_n)$ where the $F_{high}$ is replaced with a same size tensor of random numbers sampled from the normal distribution, and 3) $T_{n}^{noise} = (X_n^{noise}, Y_n)$ where the Gaussian noise is added to the images. The purpose of constructing these tasks is to test the robustness of the learned inductive bias when facing a new task with frequency disturbance or noise under the cross-domain settings and its tendency on frequency perception compared to the original meta-testing. 

From Table~\ref{MoreResultsDifferentReconstructedTargetDomainTasks}, we can learn several points. 1) We notice that for every baseline combined with our methods, the average testing accuracy results of $T_{n}^{zeros}$ and $T_{n}$ are almost similar, which means that the low-frequency do take the dominant position to represent the semantic cues on classification tasks. 2) Our method can resist the disturbance of high-frequency components, and still provide improvements when compared to all baselines. 3) TPN and GNN are more robust FSL classifiers when facing noisy samples. The former suffers the $5.4\%$ ($57.17 \to 54.09$) and $6.7\%$ ($45.31 \to 42.26$) drops of average accuracy from the baseline for 5-way 5-shot and 5-way 1-shot settings, and the latter suffers the $5.5\%$ ($56.78 \to 53.65$) and $6.4\%$ ($41.00 \to 38.38$). While the percentage numbers of RelationNet are $16.9\%$ ($50.66 \to 42.12$) and $12.4\%$ ($38.23 \to 33.49$) respectively. These observations suggest that frequency information honestly influences classification accuracy across domains.

\subsection{Analysis of Frequency-Aware Augmentation}
In this subsection, we provide the analysis of the effect and the compatibility of the Frequency-Aware Augmentation (FAA) process, which is derived from the complete proposed Frequency-Aware Prompting (FAP) method by removing the adversarial task augmentation process only. Compared with the complete FAP, the FAA has maintained the core units of FAP, including constructing the frequency-aware augment samples and employing the mutual attention modules. Exploring the effect of FAA is helpful to demonstrate the improvement of considering the CD-FSL problem from a frequency-aware perspective. Furthermore, as mentioned before, we argue that our proposed Frequency-Aware Prompting method is compatible with other CD-FSL methods. Besides combining the task-diversity improvement method, by employing the FAA to the feature-wise manipulation method, such as FWT~\cite{tseng2020FWT}, we can also explain the compatibility of the proposed frequency-aware augmentation process. 

Table~\ref{MoreResultsFAPvFAA} presents the quantitative results of different combinations by using Frequency-Aware Augmentation (FAA) on the CD-FSL benchmark datasets. From the results, we can learn that either FAP or FAA is a plug-and-play augmentation method. Not only the few-shot baselines but also the feature-wise transformation method can be combined with it. And their performance is all improved by employing the FAA.

Another interesting observation is that the combination of baseline+FWT+FAA can even perform better than baseline+ FAP in several cases, which indicates the underlying potential of employing a hybrid method in CD-FSL settings. This also demonstrates that the Frequency-Aware Prompting indeed helps few-shot baselines to generalize to other domains when there is only one source domain dataset.